\documentclass{article}

\usepackage{microtype, graphicx, subfigure, booktabs, hyperref}

\usepackage[accepted]{icml2020}

\usepackage{amssymb, amsfonts, mathtools, mathrsfs, pgfplots, pgfplotstable, amsthm, commath}
\usepgfplotslibrary{groupplots}

\pgfplotstableread{Data/FGSM_MNIST_Attack.csv}
\FGSMMNISTAttack

\pgfplotstableread{Data/FGSM_MNIST_Ratio.csv}
\FGSMMNISTRatio

\pgfplotstableread{Data/LinfPGD_MNIST_Attack.csv}
\LinfPGDMNISTAttack

\pgfplotstableread{Data/LinfPGD_MNIST_Ratio.csv}
\LinfPGDMNISTRatio

\pgfplotstableread{Data/MI_MNIST_Attack.csv}
\MIMNISTAttack

\pgfplotstableread{Data/MI_MNIST_Ratio.csv}
\MIMNISTRatio

\pgfplotstableread{Data/MNIST_Ratio_ave.csv}
\MNISTRatioave

\pgfplotstableread{Data/MNIST_GDR.csv}
\MNISTGDR

\pgfplotstableread{Data/FGSM_Fashion_Attack.csv}
\FGSMFashionAttack

\pgfplotstableread{Data/FGSM_Fashion_Ratio.csv}
\FGSMFashionRatio

\pgfplotstableread{Data/LinfPGD_Fashion_Attack.csv}
\LinfPGDFashionAttack

\pgfplotstableread{Data/LinfPGD_Fashion_Ratio.csv}
\LinfPGDFashionRatio

\pgfplotstableread{Data/MI_Fashion_Attack.csv}
\MIFashionAttack

\pgfplotstableread{Data/MI_Fashion_Ratio.csv}
\MIFashionRatio

\pgfplotstableread{Data/Fashion_Ratio_ave.csv}
\FashionRatioave

\pgfplotstableread{Data/Fashion_GDR.csv}
\FashionGDR

\begin{document}

%%%%%%%%%%%%%%%%%%%%
%%% TITLE AND AUTHORS
%%%%%%%%%%%%%%%%%%%%

\twocolumn[
\icmltitle{Evaluating Ensemble Robustness Against Adversarial Attacks}

\icmlsetsymbol{equal}{*}

\begin{icmlauthorlist}
\icmlauthor{George Adam}{UofTcs}
\icmlauthor{Romain Speciel}{UofTm}
\end{icmlauthorlist}

\icmlaffiliation{UofTcs}{Department of Computer Science, University of Toronto, Toronto, Canada}
\icmlaffiliation{UofTm}{Department of Mathematics, University of Toronto, Toronto, Canada}
%NEED TO INCLUDE FIELDS INSTITUTE

\icmlcorrespondingauthor{Romain Speciel}{romain.speciel@mail.utoronto.ca}
\icmlcorrespondingauthor{Alex Adam}{alex.adam@mail.utoronto.ca}

\icmlkeywords{ICML}

\vskip 0.3in
]

\printAffiliationsAndNotice{\icmlEqualContribution}

%%%%%%%%%%%%%%%%%%%%%%%%%%%%%%%%%%%%%%%%%%%%%%%%%%%%%%%%%%%%
%%%%%%%%%%%%%%%%%%%%%%%%%%%%%%%%%%%%%%%%%%%%%%%%%%%%%%%%%%%%
%%%%%%%%%%%%%%%%%%%%%%%%%%%%%%%%%%%%%%%%%%%%%%%%%%%%%%%%%%%%

\begin{abstract}
Adversarial examples, which are slightly perturbed inputs generated with the aim of fooling a neural network, are known to transfer between models; adversaries which are effective on one model will often fool another. This concept of \textit{transferability} poses grave security concerns as it leads to the possibility of attacking models in a black box setting, during which the internal parameters of the target model are unknown. In this paper, we seek to analyze and minimize the transferability of adversaries between models within an ensemble. To this end, we introduce a gradient based measure of how effectively an ensemble's constituent models collaborate to reduce the space of adversarial examples targeting the ensemble itself. Furthermore, we demonstrate that this measure can be utilized during training as to increase an ensemble's robustness to adversarial examples.
\end{abstract}

\section{Introduction}
Neural networks are known to be vulnerable to adversarial examples, which are slightly perturbed images that yield a misclassification. This phenomenon has been widely studied, yet the task of developing an effective defense against such attacks is of significant difficulty \cite{Carlini}. Most proposed defenses fall into two categories. The first approach is to improve the training of networks as to make them less vulnerable to adversarial examples, employing methods such as adding noise to the training set or conditioning the network on pre-perturbed inputs. The second approach turns to detection; instead of attempting to correctly classify adversaries, detection methods are content with simply flagging them. However, even detection can be quite a complicated task, as shown by \cite{Carlini}. Indeed, it seems that most detection mechanisms can be circumvented if an adversary is aware of its use.

In this paper, we strive to approach the problem using ensembles instead of single models. The idea of using ensembles to increase robustness is not novel; this has been explored by many, including \cite{Tramer1}, \cite{Kariyappa}, \cite{Pang} and \cite{Bagnall}. However, we are concerned less with the actual production of robust ensembles, but rather with the evaluation of such ensembles. More specifically, we use a gradient-based approach to evaluate how effectively the models within the ensemble collaborate to establish robustness against adversaries.

\subsection{Contributions}
The main contribution is the \textit{gradient diversity rating}, a new metric which measures the effectiveness of the collaboration amongst an ensemble's constituent models against adversarial attacks. The metric is obtained through a geometric analysis of the misalignment of the ensemble's models' gradients. We demonstrate that one can use the rating during training in order to produce ensembles with high gradient misalignment (which are therefore harder to fool). Ensembles with various gradient diversity ratings are then attacked in order to support the theoretical work by establishing a correlation between the gradient diversity rating and the ensemble's robustness to adversarial attacks.

\section{Background}
Before exploring the theoretical development of the gradient diversity rating, the main contribution of this paper, we first survey related work, and introduce important metrics which serve as a motivation to the result.

\subsection{Related Work}
Adversarial examples in the context of DNNs have come into the spotlight after Szegedy et al. showed the imperceptibility of the perturbations which could fool state-of-the-art computer vision systems \cite{szegedy_intriguing_2013}. Since then, adversarial examples have been demonstrated in many other domains, notably including speech recognition \cite{carlini_audio_2018}, and malware detection \cite{grosse_adversarial_2016}. Nevertheless, CNNs in computer vision provide a convenient domain to explore adversarial attacks and defenses due to the existence of standardized test datasets, high performing CNN models reaching human or super-human accuracy on clean data, and the marked deterioration of their performance when subjected to adversarial examples to which human vision is robust.

The intra-model and inter-model transferability of adversarial examples was investigated thoroughly by \cite{Papernot2016}. It was found that adversarial examples created for fundamentally different MNIST classification models could transfer between each other. This showed that neural networks are not special in their vulnerability, and that simply hiding model details from an attacker is bound to fail since a substitute model can be trained, whose adversarial examples are likely to transfer to the hidden model.

Transferable adversarial examples pose a security risk mainly if they are able to cause a misclassification of the same class between models. Otherwise, a defender could use the disagreement between two models with similar performance as a way of detecting these adversaries. However, a targeted attack on an ensemble of models significantly increases the same-target transfer rate.

Training for diversity amongst ensembles as a defense is a relatively new idea, but has been explored  by \cite{Pang}, \cite{Bagnall} and \cite{Kariyappa}, for example. In particular, \cite{Pang} introduces diversity in ensembles by encouraging differing prediction confidence amongst secondary outcome classes, and \cite{Kariyappa} uses the pairwise angle of the model gradients to achieve diversity on adversaries. However, none of the presented result thus far perform a rigorous geometric analysis of the space of adversarial examples, a route which we explore here.

\subsection{Metrics}
Before delving into the concept of the gradient diversity rating, we introduce the two main metrics in use as to provide some motivation to the presented result. The gradient diversity rating, as introduced in the next section, will serve as a computable, measurable and practical estimate of an ensemble's \textit{collaboration rating}, explained below.

\subsubsection{Adversarial Success}
First of the two important metrics is the \textit{adversarial success}, which corresponds the the portion of adversarial examples which successfully fool all models in the ensemble. Since the aim of gradient diversity is to ensure that all models within the ensemble cannot be fooled in the same fashion, we count the success rate of the adversarial attack as the portion of examples which yielded the \textit{same} incorrect misclassification from all of the ensemble's constituents. Formally, this can be described as follows. Given a test set $T$ of inputs $x$ with ground truth label $y_x$ such that all $x$ are correctly classified by all models $f_1,...., f_n$ in an ensembles $\mathscr{E}$, and an attack $A$ which takes inputs $x$ and perturbs them to return some input $x^*$, we define
\begin{equation}
	A(\mathscr{E})=\frac{|\{x\in T \text{ s.t. } f_1(x^*)=\dots=f_n(x^*)\neq y_x\}|}{|T|}
\end{equation}
Additionally, we can define the adversarial success of an attack against a particular model $f$, rather than an ensemble, by simply taking $\mathscr{E}=\{f\}$.

\subsubsection{Collaboration Rating}\label{collaboration}
The \textit{collaboration rating} $\text{CR}$ of $\mathscr{E}$ with respect to some attack $A$ is simply
\begin{equation}
	\text{CR}_A(\mathscr{E})=\frac{A(\mathscr{E})}{\prod_{f\in \mathscr{E}}A(f)}
\label{CR}
\end{equation}
Intuitively, if the adversaries were randomly distributed near the inputs, one would expect $\text{CR}_A(\mathscr{E})\approx 1$. A low collaboration rating means the space of adversarial examples with respect to each of the models does not largely intersect; on the other hand, if an ensemble has a high collaboration rating, this denotes some alignment in the adversarial spaces. Note that, in practice, this measure is largely ineffective in a defense setting: 
% This is ground truth, cannot optimize, created a surrogate called the gdr which we then optimize. Do not mention ineffectiveness, explain rather why we cannot optimize it
it relies upon preexisting knowledge of the nature of the attack being employed by the adversary. However, it is useful in the theoretical setting as it gives a measure of how the ensemble performs relative to the individual models, effectively measuring the collaboration of the ensemble's models, which is precisely what the gradient diversity rating aims to evaluate. Intuitively, one can think of the collaboration rating as the ground truth corresponding to the ensemble's effectiveness, which the gradient diversity rating then seeks to minimize, regardless of the employed attack.

\section{Ensemble Gradient Diversity Rating}
The following section contains the theoretical work at the foundation of the presented ideas. In particular, we derive the concept of an ensembles's Gradient Diversity Rating (GDR).
\subsection{Definitions}
Throughout, define $x\in \mathbb{R}^n$ to be an unperturbed image with ground truth $y$ (in a classification setting), $f$ to be some model outputting a predicted 
%probability distribution over some set of classes
class, $f^c$ to be the prediction confidence of class $c$ (i.e. $f(x) = c'$ where $f^{c'}(x)=\max_c f^c(x)$), and $\nabla f(x)$ to be the gradient of $f^y$ at $x$. Note this is the gradient of the model with respect to the \textit{correct} class, regardless of the model's output itself.
\subsection{Assumptions}
Since this is a gradient based approach, we naturally rely on the assumption that each of the models within the ensembles we discuss is \textit{approximately} linear. Our approach will work best if the following holds for (most) inputs $x$ and (small) perturbations $v$:
\begin{center}
	$v\cdot \nabla f(x)<0$ if and only if $f^y(x+v)<f^y(x)$
\end{center}
In other words, the prediction confidence increases in the direction of the gradient, and decreases in the negative direction of the gradient. This condition need only hold for most inputs and reasonable perturbations; the above criterion is not a categorical determinant of our approach's validity. Nonetheless, the theory developed in support of the approach does rely upon this assumption. Call this assumption the \textit{linearity assumption}.

% review/rethink how to present this linearity condition, can be achieved through regularization

\subsection{Theoretical Development}
We denote an adversary by $x^* = x+v$ for some small $v$ with $\norm{v}\leq \epsilon$ given a norm and $\epsilon>0$ (for example, in the case of MNIST, $|v|\leq 0.3$ by convention). Importantly, note that no assumptions are made on how this adversary is reached; regardless of the employed adversarial attack, one will always be able to express the adversary in the above form. Therefore, since the following reasoning does not rely upon knowledge of the attack method, it generalizes to any adversarial attack. Near $x$, if $v$ projects negatively on $\nabla f(x)$, it will reduce the projection confidence of $f^y$ (as a result of the linearity assumption), and is therefore more likely to generate an adversarial example on for $f$ at $x$. Considering now an ensemble of models $\mathscr{E}=\{f_1, \cdots, f_i\}$, we formulate the following condition:
\begin{center}
	\textit{Diversity Condition}: For all unperturbed images $x$, there does not exist a perturbation $v$ such that $v\cdot \nabla f(x)<0$ for all models $f\in \mathscr{E}$ simultaneously.
\end{center}
Indeed, if we can guarantee this condition, we ensure no perturbation can simultaneously lower the prediction confidence of all models.

The diversity condition can be rephrased in terms of the adversarial subspaces of each model at $x$. Define 
\begin{equation}
	\text{Adv}_x(f)=\{x^*=x+v\text{ s.t. } ||v||\leq \epsilon, f(x^*)\neq y\}
\end{equation}
$\text{Adv}_x(f)$ is the set of all potential adversaries at $x$ against $f$. The diversity condition strives to minimize the size of $\bigcap_{f\in\mathscr{E}}\text{Adv}_x(f)$, for all unperturbed $x$. Note that the linearity assumption allows us to claim that for most $x^*=x+v\in \text{Adv}_x(f)$, we have $v\cdot \nabla f(x)<0$, thereby establishing the link between the gradients of the models and the adversaries.

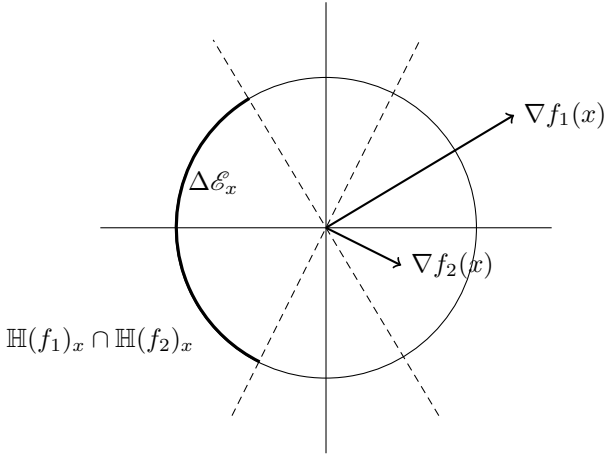
\begin{figure}[ht]
\label{grad diagram}
\begin{center}

\begin{tikzpicture}
	\draw (-3,0)--(3,0);
	\draw (0,-3)--(0,3);
	\draw (0,0) circle (2);
	\draw[->, thick] (0,0)--(2.5,1.5) node[right]{$\nabla f_1(x)$};
	\draw[->, thick] (0,0)--(1,-0.5) node[right]{$\nabla f_2(x)$};
	\draw[densely dashed] (1.5,-2.5)--(-1.5,2.5);
	\draw[densely dashed] (2.5*-0.5,2.5*-1)--(2.5*0.5,2.5*1);
	\draw[very thick] (-1.02,1.72) arc (121:243.5:2);
	\node[] at (-1.5,0.6) {$\Delta \mathscr{E}_x$};
	\node[] at (-3,-1.5) {$\mathbb{H}(f_1)_x\cap \mathbb{H}(f_2)_x$};
\end{tikzpicture}

\caption{A visualization of $\Delta \mathscr{E}_x$ in an ensemble with two models, projected onto the span of the gradients.}	

\end{center}
\end{figure}

We now progress towards a method to assess the size of the shared adversarial subspace between each of the models in the ensemble. For a given unperturbed input $x$ and model $f$, define the half space
\begin{equation}
	\mathbb{H}f_x=\{v\in \mathbb{R}^n\text{ s.t. } v\cdot \nabla f(x) < 0\}
\end{equation}
This space, which can be geometricaly described as the half space below the hyperplane naturally oriented by $\nabla f(x)$, consists of all possible perturbations that would locally reduce the ground truth prediction confidence of $f$ at $x$. Given an ensemble $\mathscr{E}$, we may now consider the convex cone
\begin{equation}
	C(\mathscr{E})_x=\bigcap_{f\in \mathscr{E}}\mathbb{H}f_x
\end{equation}
$C(\mathscr{E})_x$ contains all possible perturbations which would project negatively onto all of the gradients at $x$ of the models in the ensemble. However, only perturbations with relatively small magnitude are of interest; indeed, the perturbations must be small enough to maintain the ground truth of $x$. Therefore, we turn our attention to $C(\mathscr{E})_x$ relative to a unit sphere centered at $x$. To this end, viewing $S^{n-1}\subset \mathbb{R}^n$, define $\Delta\mathscr{E}_x= S^{n-1}\cap C(\mathscr{E})_x$. This is a spherical polytope containing all directions which simultaneously lower the prediction confidence of all models within the ensemble (see figure (\ref{grad diagram})). This points to the importance of the \textit{size} of $\Delta\mathscr{E}_x$.

In order to formalize the concept of size in this case, note first that $\Delta\mathscr{E}_x$ is an $n-1$ dimensional manifold with boundary, which can be viewed as a submanifold of $S^{n-1}$. Therefore, we establish the following rating 
\begin{equation}
	R(\mathscr{E},x)=\frac{\text{Vol}_{n-1}(\Delta\mathscr{E}_x)}{\text{Vol}_{n-1}(S^{n-1})}
\end{equation}
Where $\text{Vol}_{n-1}$ denotes $n-1$ dimensional volume, as given by the sphere's volume form. Intuitively, $R$ evaluates the portion of directions which project negatively onto all the gradients of each model in $\mathscr{E}$ at $x$. In order to obtain a more global metric of the model, we average this rating over all inputs within a test set $S$. This culminates in the definition of the ensemble's gradient diversity rating (GDR):
\begin{equation}
	\text{GDR}(\mathscr{E})=\frac{1}{|S|}\sum_{x\in S}R(\mathscr{E},x)
\label{GDR definition}
\end{equation}
A high GDR corresponds to a large portion of directions projecting negatively onto the gradients, and therefore a weaker ensemble. An optimal ensemble has $\text{GDR} = 0$. We hypothesize that this rating  directly correlates with the size of the intersection of the adversarial subspaces of all models within the ensemble, and therefore with the adversarial robustness of the ensemble.

\subsection{Analysis of the GDR}
It is worth exploring in more depth how the GDR ought to be interpreted. Indeed, it is tempting to hypothesize that, simply because some ensemble $A$ may have a lower GDR than some other ensemble $B$, ensemble $A$ will be more resistant to adversarial attacks; this is not necessarily the case. For example, consider the case when ensemble $A$ consists of three standard models trained without gradient regularization, and $B$ consists of three copies of a single model which received significant individual adversarial training. The GDR of $A$ expected to be around $\frac{1}{2^3}$, and that of $B$ is $0.5$ (See Section \ref{trivial gdr section} for the derivation of these values); however, it would not be surprising for ensemble $B$ to perform better, despite its higher GDR.

Another intriguing example is as follows: consider an ensemble consisting of as many models as output classes, with model $f_i$ always returning class $i$. While this ensemble will have consensus accuracy $0$, no adversary will be able to simultaneously fool all of the constituting models, since there will always be a model outputting the correct class. Furthermore, the GDR itself will be $0$ as no perturbation will project negatively onto all the gradients, since the gradients themselves will be $0$. However, despite the GDR being optimal, this is by no means an effective ensemble.

An ensemble's GDR is therefore better interpreted as a measure of \textit{how effectively the models within the ensemble collaborate when faced with adversarial examples}. Consequentially, the GDR must always be placed in context for it to retain significance (just as a classifier's accuracy must be placed in context of any potential class imbalance, for example). Effective collaboration is trivial when the collaborating models themselves are ineffective. Such context might include, for example, a given  ensemble $\mathscr{E}$'s GDR relative to $\frac{1}{2^{|\mathscr{E}|}}$, relative to that of another ensemble with similar architecture, or relative to the adversarial robustness of the constituent models (as in the collaboration rating, established in Section \ref{collaboration}).

\section{Calculating an Ensemble's GDR}
Evaluating an ensemble's GDR may not be as straightforward as one might initially anticipate. Indeed, evaluating $R(\mathscr{E},x)$ is analogous to evaluating a solid angle in $|\mathscr{E}|$ dimensions, a problem which is known to be hard in higher dimensions \cite{Ribando}. Therefore, we treat this problem case by case, depending on the size of the ensemble. Note that, throughout, we assume there are no linear dependences amongst gradients of the models, unless otherwise specified (this is not a costly assumption).
\subsection{$|\mathscr{E}|=1$ and other (almost) trivial cases}\label{trivial gdr section}
Trivially, when $|\mathscr{E}|=1$, we have $R(\mathscr{E},x)=0.5$ for any $x$, as there is only one gradient given, upon which exactly half of all perturbations project negatively. Therefore, in such cases, $\text{GDR}(\mathscr{E})=0.5$. Similarly, regardless of $|\mathscr{E}|$, if all models in the ensemble have the same gradient at all points, we will have $\text{GDR}(\mathscr{E})=0.5$ for the same reason. This confirms the intuition that an ensemble consisting of several copies of the same model has weak collaboration against adversaries (as one would expect), since any successful adversarial example on one model will transfer onto all others trivially. Furthermore, it is worth noting that, since any two randomly initiated models are likely to have roughly orthogonal gradients, an ensemble $\mathscr{E}$ with $|\mathscr{E}|=i$ consisting of independently trained models would be expected to have $\text{GDR}(\mathscr{E})\approx\frac{1}{2^i}$.

\subsection{$|\mathscr{E}|=2$}
Given $\mathscr{E}=\{f_1, f_2\}$ and an unperturbed image $x$, any potential perturbation $v$ can be projected first onto $\text{span}(\nabla f_1(x), \nabla f_2(x))$; from there, it can then be decided if $v$ projects negatively onto both gradients. From this, it becomes geometrically clear that 
\begin{equation}
	R(\mathscr{E},x)=\frac{1}{2\pi}\Big{(}\pi-\arccos\big{(} \nabla f_1(x)\cdot \nabla f_2(x)\big{)}\Big{)}
\end{equation}
This term is easy to compute, meaning it can be incorporated during training of the ensemble to ensure gradient diversity. Note this has already been done in \cite{Kariyappa} by minimizing the pairwise cosine similarity between gradients. However, this approach may not be optimal when $|\mathscr{E}|\geq 2$, as assumed by \cite{Kariyappa}. This is for two main reasons. First, this approach forces the gradients towards a unique geometrical layout (an $|\mathscr{E}|-1$ dimensional regular simplex), which adds unnecessary geometrical complications during training. Second, as $|\mathscr{E}|$ gets larger, this approach becomes equivalent to demanding that the gradients be pairwise orthogonal which, as previously discussed, is equivalent to simply training the models individually, with no gradient term.

\subsection{$|\mathscr{E}|=3$}
As with the previous case, we begin by projecting any potential perturbation onto $\text{span}(\nabla f_1(x), \nabla f_2(x), \nabla f_3(x))$. This case then reduces to finding the (2-dimensional) volume a spherical triangle. Alternatively, one can observe that the optimal gradient layout is reached if and only if the sum of the pairwise angles between the gradients is $2\pi$. Either of these approaches can be utilized to train an ensemble of three models to ensure gradient diversity.

\subsection{$|\mathscr{E}|=4$}
This case is treated in an almost identical method as the previous one. Indeed, after projecting on the space spanned by each of the gradients, one notes that this problem is equivalent to finding the volume of a spherical tetrahedron, the volume of which is hard to evaluate directly. Therefore, just as it was possible to find the sum of the pairwise angles between gradients when $|\mathscr{E}|=3$, we can perceive this problem as maximizing the sum of the areas of the spherical triangles between every three of the four gradients. Geometrically, it becomes clear that the maximum is reached when this sum is $4\pi$, the surface area of a sphere. This approach is much more computationally feasible, and can therefore be incorporated during training.

\subsection{$|\mathscr{E}|\geq 5$}
When $|\mathscr{E}|\geq 5$, $R(\mathscr{E},x)$ becomes hard to evaluate as there is no direct method of computing the volume of higher dimensional spherical polytopes (on spheres $S^n$ with $n\geq 3$). Indeed, the best method to obtain an exact measure of a higher dimensional solid angle calls upon a multivariate Taylor Series \cite{Ribando}. In order to circumvent this problem, we can re-express the function $R$ as
\begin{equation}
	R(\mathscr{E},x)=\int_{S^{n-1}}\prod_{f\in \mathscr{E}} H(-v\cdot \nabla f(x))\text{ d}v
\end{equation}
 with $H$ denoting the Heaviside step function and the integral ranging over the hypersphere, with the integrand valued at $1$ if $v\in S^{n-1}$ projects negatively onto each of the gradients simultaneously, and 0 otherwise. The fact that this does indeed yield the value of $R$ is immediate. The advantage of viewing the problem in such a way is that it may now be estimated with a simple Monte Carlo approach, regardless of the size of the ensemble (generating a uniform random distribution on the $n$-sphere requires a quick trick \cite{Tian}). Averaging these estimations over a test set will give an approximation to the ensemble's GDR. While this is unlikely to be useful during training, it remains an important metric of any ensemble.

Alternatively, it is indirectly demonstrated by \cite{Kariyappa} that simpler training methods can be utilized to minimize the GDR without directly computing it. This yields an import circumvention the above problem, and allows for gradient training in ensembles of arbitrary size (although the gradient loss terms may not be optimal).
\section{Ensemble Training with GDR Minimization}
As previously mentioned, one can use an ensemble's GDR during training to reduce the size of the adversarial space effective on all models simultaneously. Implementations of such methods usually will train all models in the ensemble simultaneously, utilizing a loss function of the following form:
\begin{equation}
	\text{Loss} = \text{Image Loss} + \beta(\text{Gradient Loss})
\end{equation}
The image loss term depends simply on the accuracy of the ensemble (i.e. average cross entropy loss over constituent models) and the gradient loss term consists of some method, such as those outlined in the previous section, which strives to minimize the ensemble's GDR. $\beta$ is a hyperparameter which varies upon the previous two terms, and which is best determined through experimentation. In this section, the above method is employed to create several ensembles with varying GDR, which are then tested against various attacks. The aim of this experiment is to demonstrate a correlation between an ensemble's GDR and the robustness of the ensemble to adversarial attacks.

\subsection{Experimental Design}\label{exp setup}
In order to train ensembles with varying GDR, we employed the aforementioned method. This then allows us to carry attacks on a wide range of ensembles and compare their resistance with respect to their GDR. 

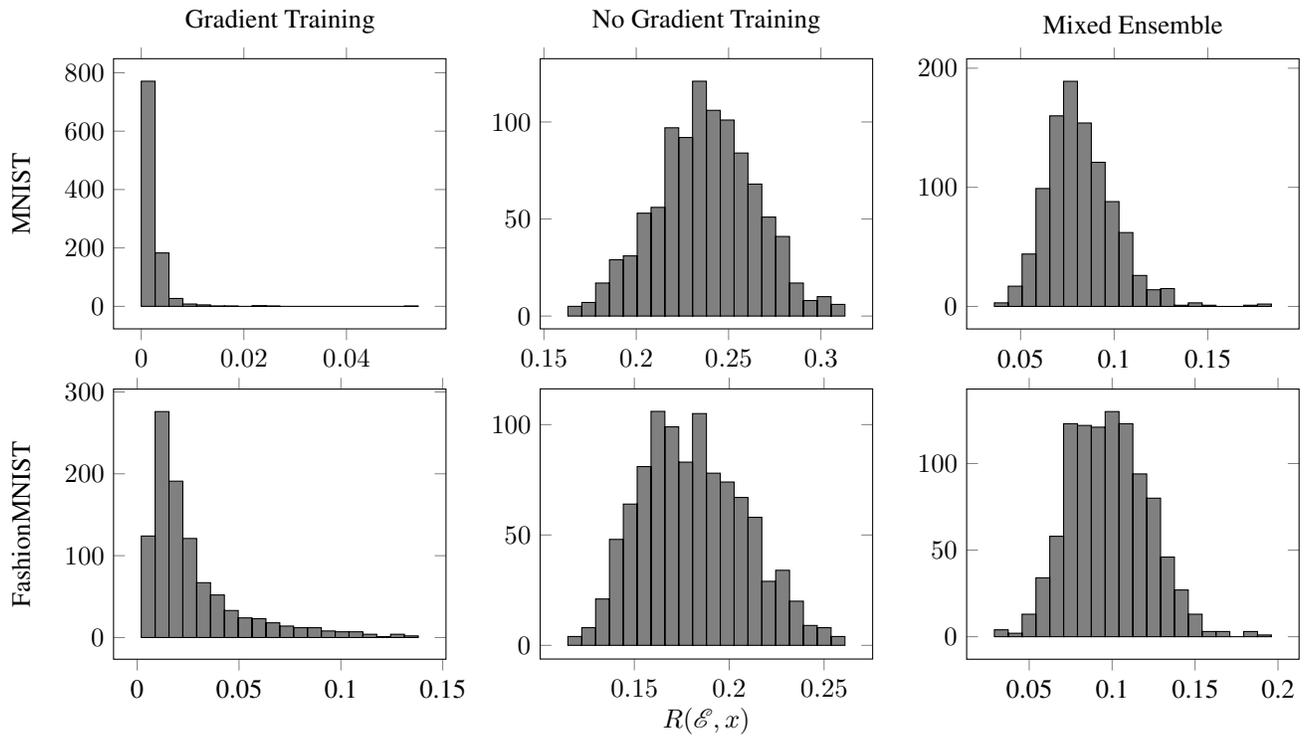
\begin{figure*}[ht]
\centering
\begin{tikzpicture}
	\begin{groupplot}[group style={group size=3 by 3, horizontal sep=1.25cm, vertical sep=0.8cm},width=0.35*\textwidth]

    \nextgroupplot[
    title = Gradient Training,
    ybar,
	ylabel={MNIST},
    xticklabel style={
  	/pgf/number format/precision=3,
  	/pgf/number format/fixed},
 	scaled x ticks=false]
	\addplot [fill=gray, hist={bins=20}] table[y=a] from \MNISTGDR;

    \nextgroupplot[
    title = No Gradient Training,
    ybar]
	\addplot [fill=gray, hist={bins=20}] table[y=b] from \MNISTGDR;

    \nextgroupplot[
    title = Mixed Ensemble,
    ybar,
	xtick={0, 0.05, 0.1, 0.15},
    xticklabels={0, 0.05, 0.1, 0.15}]
	\addplot [fill=gray, hist={bins=20}] table[y=c] from \MNISTGDR;

	\nextgroupplot[
    ybar,
	ylabel={FashionMNIST},
	xtick={0, 0.05, 0.1, 0.15},
    xticklabels={0, 0.05, 0.1, 0.15}]
	\addplot [fill=gray, hist={bins=20}] table[y=a] from \FashionGDR;

	\nextgroupplot[
    ybar, xlabel={$R(\mathscr{E},x)$}]
	\addplot [fill=gray, hist={bins=20}] table[y=b] from \FashionGDR;

	\nextgroupplot[
    ybar,
	xtick={0.05, 0.1, 0.15, 0.2},
    xticklabels={0.05, 0.1, 0.15, 0.2}]
	\addplot [fill=gray, hist={bins=20}] table[y=c] from \FashionGDR;

	\end{groupplot}
\end{tikzpicture}
\caption{Histogram of $R(\mathscr{E},x)$ for 1000 images on 3 different ensembles: one with gradient training, one without, and one consisting of two models from an ensemble with gradient training, and a third from one without. The first row consists of the MNIST ensembles, and the second of the FashionMNIST ensembles. This demonstrates that the GDR training methods are effectively reducing the GDR (see equation (\ref{GDR definition})).}
\label{GDR}
\end{figure*}

On MNIST, we trained 5 ensembles, each consisting of three individual models. The first three ensembles are trained for 15 epochs with a gradient loss term which minimizes the maximum pairwise cosine similarity of the gradients, then 15 epochs with a gradient loss term which maximizes the sum of the pairwise angles of the gradients. The final two ensembles are trained for 30 epochs without any gradient regularization. Additional ensembles are formed by recombining the individual models from these five original ensembles. This exact process is replicated on FashionMNIST. The effectiveness of this training is apparent from figure (\ref{GDR}). All ensembles reached ensemble consensus accuracy $>0.98$ and $>0.82$ for MNIST and FashionMNIST, respectively. The ensembles are subjected to three white-box attacks: the Fast Gradient Sign Method (FGSM), the $L_{\inf}$ Projective Gradient Descent (LinfPGD) and the Momentum Iterative (MI) attacks.

%\textbf{Coming soon}:  a wider variety of white box attacks, on a wider variety of data sets (MNIST, FashionMNIST, CIFAR-10...), with larger ensembles

\subsection{Results}
The results of the above experiment are displayed in figure (3). The first clear observation is that ensembles with lower GDR do indeed tend to have lower attack success and collaboration ratings, as predicted by the theoretical development of the metric. The LinfPGD attack was very effective, both against individual models and ensembles; as a result, the trends are much more apparent for low $\epsilon$ values as high $\epsilon$ values resulted in high attack success rates, regardless of the gradient training. Furthermore, due to limits in computing power, both the LinfPGD and MI attacks were conducted on a mere 500 images (in contrast to the 10,000 images which we perturbed by FGSM). This may explain the high variance in the latter two attacks.

\section{Conclusion}
The primary result of this paper is the gradient diversity rating (GDR). In particular, the GDR metric provides a direct indication of the collaboration strength between to constituent models of an ensemble, and can demonstrably be used during training to create more robust ensembles. Early experimentation suggests that, as expected, an ensemble with a lower GDR is less vulnerable to attacks fooling all of its constituting models. However, there is much room left for improvements and clarification. Interesting topics to explore could include the analysis of GDR on ensembles consisting of models with individual adversarial training, methods to improve attacks against ensembles with gradient diversity training, and improvements to the GDR metric to yield a more absolute, comparable measure of the ensemble's effectiveness, perhaps by introducing terms related to model accuracy or number of output classes.

%%%%%%%%%%%%%%%%%%%%%%%%%%%%%%%%%%%%%%%%%%%%%%%%%%%%%%%%%%%%

%%%%%%%%%%%%%%%%%%%%
%%% Bibliography
%%%%%%%%%%%%%%%%%%%%

\newpage
\bibliography{biblio}
\bibliographystyle{icml2020}

%%%%%%%%%%%%%%%%%%%%%%%%%%%%%%%%%%%%%%%%%%%%%%%%%%%%%%%%%%%%

%%%%%%%%%%%%%%%%%%%%
%%% Figures
%%%%%%%%%%%%%%%%%%%%

%%%%%%%%%%%%%%%%%%%%

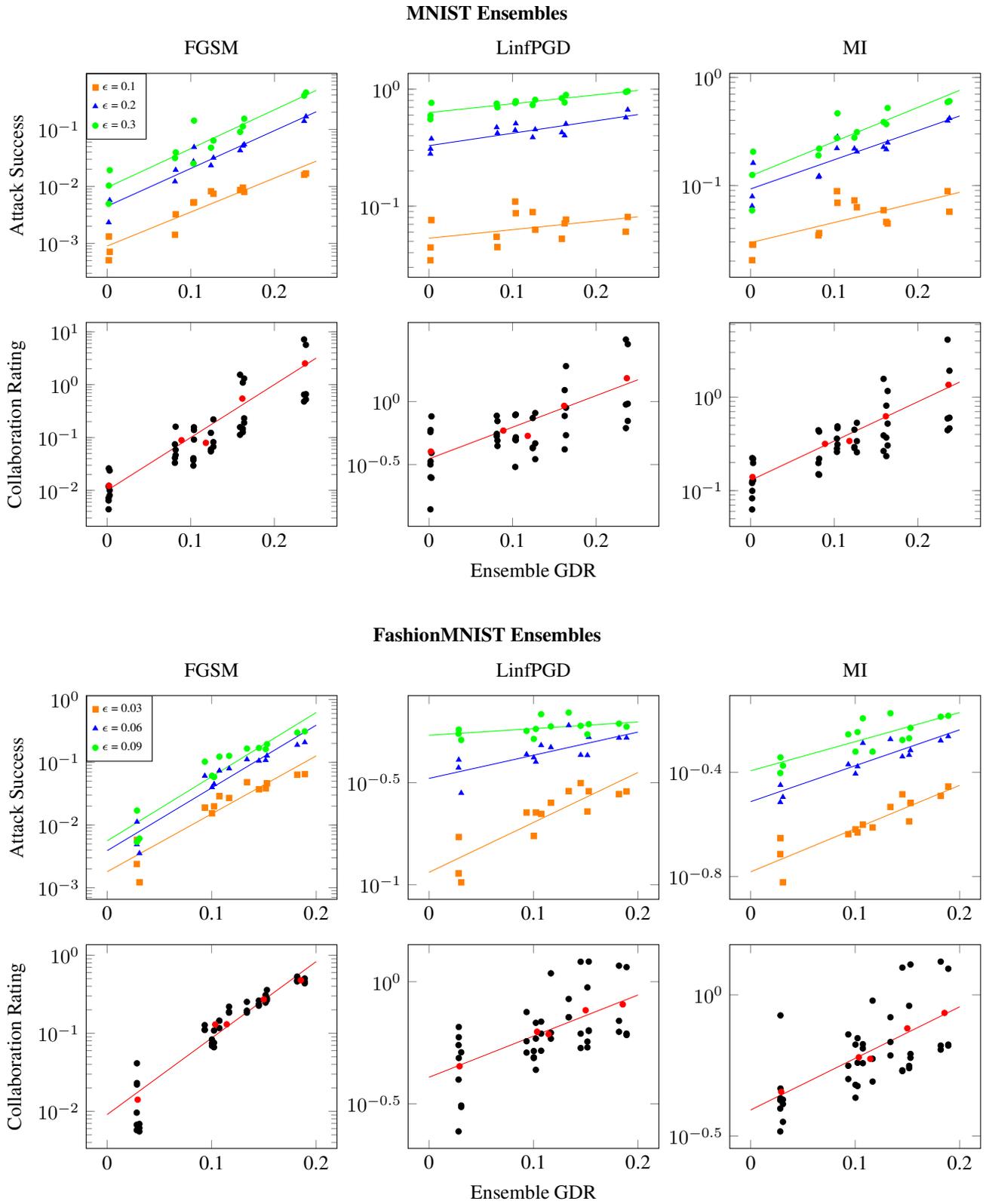
\begin{figure*}[ht]\label{results}
\centering
% MNIST Results

\textbf{MNIST Ensembles}\par\medskip
\begin{tikzpicture}
  \begin{groupplot}[group style={group size=3 by 2, horizontal sep=1.25cm, vertical sep=0.8cm},width=0.35*\textwidth]

    \nextgroupplot[
    title = FGSM,
    ylabel= Attack Success,
    ymode=log,
    xtick={0, 0.1, 0.2},
    xticklabels={0, 0.1, 0.2},
    legend style={at={(0,1)},anchor=north west, font = \tiny}]
    \addplot[only marks, mark options={scale=0.7}, mark=square*, color=orange] table[y = a] from \FGSMMNISTAttack;
	\addlegendentry{$\epsilon$ = 0.1};
	\addlegendentry{$\epsilon$ = 0.2};
	\addlegendentry{$\epsilon$ = 0.3};
	\addplot[only marks, mark options={scale=0.7}, mark=triangle*, color=blue] table[y = b] from \FGSMMNISTAttack;
	\addplot[only marks, mark options={scale=0.7}, color=green] table[y = c] from \FGSMMNISTAttack;
	\addplot [domain=0:0.25, color=orange] {0.0009 * (e^(13.717 * x))};
	\addplot [domain=0:0.25, color=blue] {0.0045 * (e^(15.248 * x))};
	\addplot [domain=0:0.25, color=green] {0.0096 * (e^(15.669 * x))};

	\nextgroupplot[
	title = LinfPGD,
    ymode=log,
   	xtick={0, 0.1, 0.2},
    xticklabels={0, 0.1, 0.2}]
	\addplot[only marks, mark options={scale=0.7}, mark=square*, color=orange] table[y = a] from \LinfPGDMNISTAttack;
	\addplot[only marks, mark options={scale=0.7}, mark=triangle*, color=blue] table[y = b] from \LinfPGDMNISTAttack;
	\addplot[only marks, mark options={scale=0.7}, color=green] table[y = c] from \LinfPGDMNISTAttack;
	
	\addplot [domain=0:0.25, color=orange] {0.0532 * (e^(1.6846 * x))};
	\addplot [domain=0:0.25, color=blue] {0.3291 * (e^(2.4425 * x))};
	\addplot [domain=0:0.25, color=green] {0.6298 * (e^(1.7489 * x))};

	\nextgroupplot[
	title = MI,
    ymode=log,
    xtick={0, 0.1, 0.2},
    xticklabels={0, 0.1, 0.2}]
	\addplot[only marks, mark options={scale=0.7}, mark=square*, color=orange] table[y = a] from \MIMNISTAttack;
	\addplot[only marks, mark options={scale=0.7}, mark=triangle*, color=blue] table[y = b] from \MIMNISTAttack;
	\addplot[only marks, mark options={scale=0.7}, color=green] table[y = c] from \MIMNISTAttack;
	\addplot [domain=0:0.25, color=orange] {0.0296 * (e^(4.2875 * x))};
	\addplot [domain=0:0.25, color=blue] {0.093 * (e^(6.2123 * x))};
	\addplot [domain=0:0.25, color=green] {0.1227 * (e^(7.2876 * x))};

	\nextgroupplot[
	ylabel=Collaboration Rating,
    ymode=log,
    xtick={0, 0.1, 0.2},
    xticklabels={0, 0.1, 0.2}]
    \addplot[only marks, mark options={scale=0.7}] table[y = a] from \FGSMMNISTRatio;
	\addplot[only marks, mark options={scale=0.7}] table[y = b] from \FGSMMNISTRatio;
	\addplot[only marks, mark options={scale=0.7}] table[y = c] from \FGSMMNISTRatio;  
	\addplot[only marks, mark options={scale=0.7}, color=red] table[y = a] from \MNISTRatioave;
	\addplot [domain=0:0.25, color=red] {0.01 * (e^(23.034 * x))}; 
	 
	\nextgroupplot[
	xlabel= Ensemble GDR,
    ymode=log,
    xtick={0, 0.1, 0.2},
    xticklabels={0, 0.1, 0.2},
    ytick ={0.316,1}]
    \addplot[only marks, mark options={scale=0.7}] table[y = a] from \LinfPGDMNISTRatio;
	\addplot[only marks, mark options={scale=0.7}] table[y = b] from \LinfPGDMNISTRatio;
	\addplot[only marks, mark options={scale=0.7}] table[y = c] from \LinfPGDMNISTRatio;
	\addplot[only marks, mark options={scale=0.7}, color=red] table[y = b] from \MNISTRatioave;
	\addplot [domain=0:0.25, color=red] {0.3508 * (e^(5.7851 * x))};   
	
	\nextgroupplot[
    ymode=log,
    xtick={0, 0.1, 0.2},
    xticklabels={0, 0.1, 0.2}]
    \addplot[only marks, mark options={scale=0.7}] table[y = a] from \MIMNISTRatio;
	\addplot[only marks, mark options={scale=0.7}] table[y = b] from \MIMNISTRatio;
	\addplot[only marks, mark options={scale=0.7}] table[y = c] from \MIMNISTRatio; 
	\addplot[only marks, mark options={scale=0.7}, color=red] table[y = c] from \MNISTRatioave;
	\addplot [domain=0:0.25, color=red] {0.1291 * (e^(9.6665 * x))};
  \end{groupplot}
\end{tikzpicture}
\vspace{0.5em}

% FashionMNIST results
\textbf{FashionMNIST Ensembles}\par\medskip
\begin{tikzpicture}
  \begin{groupplot}[group style={group size=3 by 2, horizontal sep=1.25cm, vertical sep=0.8cm},width=0.35*\textwidth]

    \nextgroupplot[
    title = FGSM,
    ylabel= Attack Success,
    ymode=log,
    xtick={0, 0.1, 0.2},
    xticklabels={0, 0.1, 0.2},
    legend style={at={(0,1)},anchor=north west, font = \tiny}]
    \addplot[only marks, mark options={scale=0.7}, mark=square*, color=orange] table[y = a] from \FGSMFashionAttack;
	\addlegendentry{$\epsilon$ = 0.03};
	\addlegendentry{$\epsilon$ = 0.06};
	\addlegendentry{$\epsilon$ = 0.09};
	\addplot[only marks, mark options={scale=0.7}, mark=triangle*, color=blue] table[y = b] from \FGSMFashionAttack;
	\addplot[only marks, mark options={scale=0.7}, color=green] table[y = c] from \FGSMFashionAttack;
	\addplot [domain=0:0.2, color=orange] {0.0018 * (e^(21.255 * x))};
	\addplot [domain=0:0.2, color=blue] {0.0039 * (e^(23.037 * x))};
	\addplot [domain=0:0.2, color=green] {0.0056 * (e^(23.551 * x))};

	\nextgroupplot[
	title = LinfPGD,
    ymode=log,
   	xtick={0, 0.1, 0.2},
    xticklabels={0, 0.1, 0.2}]
	\addplot[only marks, mark options={scale=0.7}, mark=square*, color=orange] table[y = a] from \LinfPGDFashionAttack;
	\addplot[only marks, mark options={scale=0.7}, mark=triangle*, color=blue] table[y = b] from \LinfPGDFashionAttack;
	\addplot[only marks, mark options={scale=0.7}, color=green] table[y = c] from \LinfPGDFashionAttack;
	
	\addplot [domain=0:0.2, color=orange] {0.1149 * (e^(5.6415 * x))};
	\addplot [domain=0:0.2, color=blue] {0.3321 * (e^(2.6175 * x))};
	\addplot [domain=0:0.2, color=green] {0.5408 * (e^(0.7497 * x))};

	\nextgroupplot[
	title = MI,
    ymode=log,
    xtick={0, 0.1, 0.2},
    xticklabels={0, 0.1, 0.2},
    ytick={0.1585,0.3981}]
	\addplot[only marks, mark options={scale=0.7}, mark=square*, color=orange] table[y = a] from \MIFashionAttack;
	\addplot[only marks, mark options={scale=0.7}, mark=triangle*, color=blue] table[y = b] from \MIFashionAttack;
	\addplot[only marks, mark options={scale=0.7}, color=green] table[y = c] from \MIFashionAttack;
	\addplot [domain=0:0.2, color=orange] {0.1652 * (e^(3.8236 * x))};
	\addplot [domain=0:0.2, color=blue] {0.3073 * (e^(3.174 * x))};
	\addplot [domain=0:0.2, color=green] {0.4035 * (e^(2.5775 * x))};

	\nextgroupplot[
	ylabel=Collaboration Rating,
    ymode=log,
    xtick={0, 0.1, 0.2},
    xticklabels={0, 0.1, 0.2}]
    \addplot[only marks, mark options={scale=0.7}] table[y = a] from \FGSMFashionRatio;
	\addplot[only marks, mark options={scale=0.7}] table[y = b] from \FGSMFashionRatio;
	\addplot[only marks, mark options={scale=0.7}] table[y = c] from \FGSMFashionRatio;
	\addplot[only marks, mark options={scale=0.7}, color=red] table[y = a] from \FashionRatioave;
	\addplot [domain=0:0.2, color=red] {0.0091 * (e^(22.552 * x))};  
	 
	\nextgroupplot[
	xlabel= Ensemble GDR,
    ymode=log,
    xtick={0, 0.1, 0.2},
    xticklabels={0, 0.1, 0.2},
    ytick ={0.316,1}]
    \addplot[only marks, mark options={scale=0.7}] table[y = a] from \LinfPGDFashionRatio;
	\addplot[only marks, mark options={scale=0.7}] table[y = b] from \LinfPGDFashionRatio;
	\addplot[only marks, mark options={scale=0.7}] table[y = c] from \LinfPGDFashionRatio;
	\addplot[only marks, mark options={scale=0.7}, color=red] table[y = b] from \FashionRatioave;
	\addplot [domain=0:0.2, color=red] {0.406 * (e^(3.8807 * x))};  
	
	\nextgroupplot[
    ymode=log,
    xtick={0, 0.1, 0.2},
    xticklabels={0, 0.1, 0.2}]
    \addplot[only marks, mark options={scale=0.7}] table[y = a] from \MIFashionRatio;
	\addplot[only marks, mark options={scale=0.7}] table[y = b] from \MIFashionRatio;
	\addplot[only marks, mark options={scale=0.7}] table[y = c] from \MIFashionRatio;
	\addplot[only marks, mark options={scale=0.7}, color=red] table[y = c] from \FashionRatioave;
	\addplot [domain=0:0.2, color=red] {0.3912 * (e^(4.2027 * x))}; 
  \end{groupplot}
\end{tikzpicture}

\caption{Performance of the MNIST (top) and FashionMNIST (bottom) ensembles against three attacks (FGSM, LinfPGD, MI). The attack success corresponds to the portion of adversaries which successfully fooled the ensembles. The collaboration rating denotes the performance of the model relative to its constituent models, as defined by equation (\ref{CR}). The red points denote the average of the ensembles with the same composition.  The ensembles' GDR is recorded along the $x$-axis. Note the overall positive trend, which suggest that a low GDR does indeed result in a more robust ensemble, as predicted by the theoretical development of the metric.}
\end{figure*}

\end{document}